Analyzing the Variations in Emergency Department Boarding and Testing the Transferability of Forecasting Models across COVID-19 Pandemic Waves in Hong Kong: Hybrid CNN-LSTM approach to quantifying building-level socioecological risk

Eman Leung[1#], Jingjing Guan[1#], Kin On Kwok[1], CT Hung[1], CC. Ching[1], CK. Chung[1], Hector Tsang[2], EK Yeoh[1], Albert Lee[1*]

1 JC School of Public Health and Primary Care, The Chinese University of Hong Kong

2 Department of Rehabilitation Science, Hong Kong Polytechnic University

# first author

* corresponding author

**Title**

Analyzing the Variations in Emergency Department Boarding and Testing the Transferability of Forecasting Models across COVID-19 Pandemic Waves in Hong Kong: Hybrid CNN-LSTM approach to quantifying building-level socioecological risk

**Abstract**

Emergency department's (ED) boarding (defined as ED waiting time greater than four hours) has been linked to poor patient outcomes and health system performance. However, little is known about the extent of ED boarding across different epidemic waves of COVID-19. Nor have the complex and dynamic relationships among past ED boarding, the time-varying COVID-19 case counts and the corresponding public health and social measures, the time invariant risk factors exit in the socioecology of the catchment population, considered in previous forecasting models of ED boarding.

Hence, to forecast ED boarding across different waves of COVID-19 pandemic in the context of a complex and dynamic regional medical system in HK, a hybrid convolutional neural network (CNN)-Long short-term memory (LSTM) model was applied to public-domain data sourced from Hong Kong's Hospital Authority, Department of Health, and Housing Authority. In addition, we sought to identify the phase of the COVID-19 pandemic that most significantly perturbed our complex adaptive healthcare system, thereby revealing a stable pattern of interconnectedness among its components. To achieve this, we constructed models from each phase and tested their performance against data from all other phases. This deep transfer learning approach allowed us to determine which phase's model could most consistently explain the patterns observed in all other phases, beyond the model built from indigenous data.

Of model built from the wave or period showing the greatest proportion of days with surges in ED attendance to data extracted from other waves or period of the pandemic. It is expected that even when transferred to data extracted from when the system was less perturbed, the model built from when the system was most perturbed outperformed, and added value to, the indigenous models built from data extracted from a less perturbed wave or period.

Our result shows that 1) the greatest proportion of days with ED boarding was found between waves four and five; 2) the best-performing model for forecasting ED boarding was observed between waves four and five, which was based on features representing time-invariant residential buildings' built environment and sociodemographic profiles and the historical time series of ED boarding and case counts, compared to during the waves when best-performing forecasting is based on time-series features alone; and 3) when the model built from the period between waves four and five was applied to data from other waves via deep transfer learning, the transferred model enhanced the performance of indigenous models.

In conclusion, our research underscores the impact of residential built environment and sociodemographic profiles on ED boarding during the period between the fourth and fifth pandemic waves. This critical period, marked by perturbation to the healthcare system, exposed the interrelatedness of its various elements. As we are now navigating a similar phase of coexistence with the virus, our findings could guide resource distribution and enhance readiness for impending outbreaks.

**Introduction**

Emergency department (ED) crowding is a well-known threat to patient safety(1), and the documented adverse effects range from a decrease in the work satisfaction of ED staff (2) to increased length of stay (3) and increased mortality (4-7). In addition, EDs are confronted with limited resources (human and material) in the face of a growing demand for emergency medical care(8-9). Hence, it was noted that EDs must integrate into their mode of operation a sufficient level of resilience (or "proactive capacity") that allows them to anticipate demands associated with care needs and quickly mobilize the necessary resources to meet the demand for care(10-12).

However, anticipating patient demand in ED is challenging compared to other hospital departments for the following reasons. Firstly, being the first access point for patients who suffer from care needs that are urgent and accidental (13), ED's serve patients whose needs are more diverse, treatment level more varied, and their arrival time unexpected than other hospital departments (14). Secondly, in contrast to outpatient clinics or elective surgery, EDs are unable to adjust the inflow of patients, and are thus exposed to both a stochastic incidence of diseases and changes in patients' care-seeking behavior, nor are ED able to freely adjust the outflow of patients as it depends on other healthcare facilities to organize follow-up care.

Hence, accurate forecasting (hourly or daily) of patient demands for ED care is a key enabler to a proactive patient flow strategy for the better use of available resources and avoid overcrowding that may lead to strain situations(15-19). Over the years, several methods have been developed to improve the accuracy of modeling and forecasting ED demands. These methods include ARIMA and discrete event simulation models. A few studies applied them to predict short-term crowding specifically based on recent history and seasonality (20-22).

While each method has its own shortcoming (23), ARIMA and its variants are the most widely used methods in ED forecasting studies (24). To date, these studies have largely focused on predicting patient arrival and departure volumes with limited clinical data (25-28) or patient-level outcomes based on electronic medical record (EMR) data (29-32) with univariate ARIMA models (or its variants/extensions) constructed from recent history or seasonality (Smith et al.(33)) for a brief review). On the other hand, crowding was only an infrequent topic of focus and a poorly defined variable to forecast in these studies (33). In addition, while the intention to forecast crowding is to inform preemptive mitigation strategies, the univariate nature of these models may preclude our ability to identify intervention targets among a complex array of interconnected elements in our healthcare system that may contribute to crowding (23-24, 33). Most notably, forecasting with ARIMA model (or its variants/extensions) can only reach a satisfactory performance if patients' arrival time-series data exhibits regular variations (24). Hence, perturbations to the healthcare system, such as COVID-19, caused irregular variations in the time-series data of ED crowding as a reflection of multivariate contribution over time, challenging traditional forecasting models.

To address the irregular variations in time-series data and to consider a complex array of interconnected contributors at the population or health system level, machine learning (34-39) and deep learning (24) models have been applied to study a broad spectrum of healthcare issues, ranging from forecasting ED visits to COVID-19 modeling and forecasting. More recently, partly in response to the COVID-19 pandemic, several studies have developed machine learning and deep learning models to forecast the population-level spread of the virus, (40-43) and to predict individual patient outcomes and pathways through the health system (e.g., length of stay, bed occupancy) during COVID-19. (44-46) However, although the findings of these studies have implications on crowding at the ED, and these methods can be applied in future research to estimate the likelihood of ED crowding in the context of COVID-19 or even other pandemics, ED crowding was not the focus of the aforementioned studies. There remains a gap in research predicting ED crowding in the context of COVID-19 (33). A notable exception is the work of Smith et al. (33). Using a machine learning model, Smith et al. predicted ED crowding during the pre-COVID period and at the beginning of COVID -19 in terms of the proportion of ED boarding (that is ED LOS >= 4hr; both NHS (47-51) and the Joint Commission (52) had benchmarked ED waiting time at less than 4 hours given the evidence on patient safety). In addition, Smith et al. also examined the data drift when the model was applied across the pre- and peri-COVID periods to demonstrate whether the model is transferable over time, and the relationship among the different interconnected elements of the healthcare system studied remains the same despite the perturbation brought about by COVID-19.

The current study adds to value this emerging body of literature by 1) forecasting with a deep learning model, the ED boarding observed at different waves of the pandemic (which is defined by different dominant variants of the virus whose transmissibility and the associated social and public health measures were different (53)) and the period between waves four and five when the population was living with the virus had laid domant but only to resurge subsequently; 2) applying deep transfer learning methodology to examine the performance of models built from data of each wave (or the period) when transferred to data extracted from other waves or periods and compared with that of the respective indigonous models; and 3) quantifying the relative importance of features representing the studied ED's neighboring residents' internal and external built environment, sociodemographics, and historical COVID-19 case count with respect to the history of boarding at the ED studied. This knowledge will facilitate improved preparedness of the healthcare system for upcoming endemic scenarios (54), as we strive to coexist with the virus.

**Method**

*Data Sources*

Data were extracted from multiple public sources. The Health Bureau's Hospital Authority provided hourly records of the emergency department (ED) waiting time for the studied hospital, while the Department of Health made available the accumulated COVID-19 case counts for each building in the studied district throughout the pandemic. ED waiting time records were extracted for the period between December 31, 2018, and July 27, 2022. Daily COVID-19 confirmed cases in buildings within the catchment area of the hospital were extracted separately for Wave One to Four (January 23, 2020, to May 21, 2021), the period between Waves Four and Five (May 22, 2021, to December 23, 2021), and Wave Five (December 24, 2021, to July 23, 2022). Additionally, data related to the sociodemographic characteristics of residential buildings, as well as their internal (architectural) and external built environment, were obtained from the Census and

Statistics Department, Housing Authority, and Google Map, respectively. It is important to note that no changes in residential built environment or sociodemographic were observed during the study period. Please refer to Table 1 for a comprehensive list of residential built environment and sociodemographic features along with the descriptive statistics.

*Analytic Models*

We constructed and validated a hybrid CNN-LSTM model (55) in a sequential manner, utilizing a rolling 7-day input with a 14-day outcome for each period. These periods included individuals waves of the pandemic as well as the interval between waves four and five. While the benefit of combining CNN's sensitivity to local patterns and LSTM's supremacy over RNN in learning long-term trend with more parameters in forecasting with timeseries data had been recently recognized (55), the application of the hybrid CNN-LSTM model to handle simultaneously timeseries input and time-invariant input had not been explored. Moreover, our approach to organizing the time-invariant input features to reflect the socioecology of the district was innovative. Each time-invariant input layer consisted of all levels of a single feature, while different layers representing features from the same socioecological levels (building-level, estate-level, and TPU-level) were concatenated and served as input to a convolutional layer.

Notably, deep learning models enable the estimation of the unique contribution of each input feature while considering the contributions of all other features and their potential interactions. Nonetheless, the hierarchically organized concatenation of sociodemographic and environmental features also allows the importance associated with interaction among features from the same or different input layers to be estimated in isolation within the context of other features' contributions, singly or in combination. In order to accurately determine the relative contributions of historical surge patterns, case counts, and sociodemographic and environmental factors, while accounting for the former two, this study compared the performance of different models in each period. These models included: 1) only the historical patterns of breaching the 4-hour target, 2) both the historical patterns and timeseries of COVID-19 case counts, and 3) historical patterns, case counts timeseries, as well as residential built environment and sociodemographic profiles.

*SHAP (SHapley Additive exPlanations)*

A SHAP Explainer (56-57) was deployed to quantify the importance (as parameterized as Shapely value) of those sociodemographic or environmental features contributed to the forecasting of surges in ED attendance over a 14-day horizon. Shapely value is a method based on cooperative game theory and has been applied to increase the transparency and interpretability of machine learning models (58). Shapely value is a numerical expression of the marginal contribution of each parameter to the outcome (i.e. feature importance). The unique contribution of each parameter can be expressed as the degree of change in overall performance when the parameter is excluded. Shapely value is more sensitive, consistent, and accurate than the standardized regression coefficient, which is also aimed at parameterizing unique contributions of individual features, but does so in linear models (34). Notably, SHAP has been dubbed as the key tool in the explainable artificial intelligence's approach to making deep learning models more interpretable to end-users (34). In the current study, a Shapely value was assigned to every response level of each feature in table M1. The Shapely values were obtained by using the Python library *Shap* (59).

*Transfer Learning*

Machine learning algorithms are traditionally designed to train every model in isolation based on the specific domain, data, and task to solve specific tasks. The models have to be rebuilt from scratch once the feature-space distribution changes. Transfer learning is a means to overcome the isolated learning paradigm and generalize the knowledge acquired for one task to solve *related* ones (61). For this reason, the extent to which a model could be transferred across domains and data reflects the relatedness of the domains and/or data in question. And it is the *transferability* of the learning model built between waves four and five that the current study sought to test.

The machinic of testing the transferability of models with transfer learning is straightforward. In machine learning, a large amount of training data is typically required for training a model from scratch. Transfer learning generalizes the machine learning model developed from one context (domain) to another despite large amounts of training data or data points that are common across the two domains which are not available. Specifically, transfer learning enables us to utilize knowledge from previously learned tasks and apply them to newer, related ones. If we have significantly more data for task T1, we may utilize its learnings and generalize them for task T2 despite T2 having significantly fewer data. Compared to the ignorant learner model, transfer learning can improve baseline performance, reduce the overall amount of time taken to develop/learn, and improve the final performance of a model. Transfer learning creates an inductive bias on how and what is learned by the new model in performing a similar task within the context of a different domain. It does so by narrowing the hypothesis space from which the new model is developed, and facilitating the process with which learning is performed within the hypothesis space (for a schematic illustration, please refer to Figure M1 below).

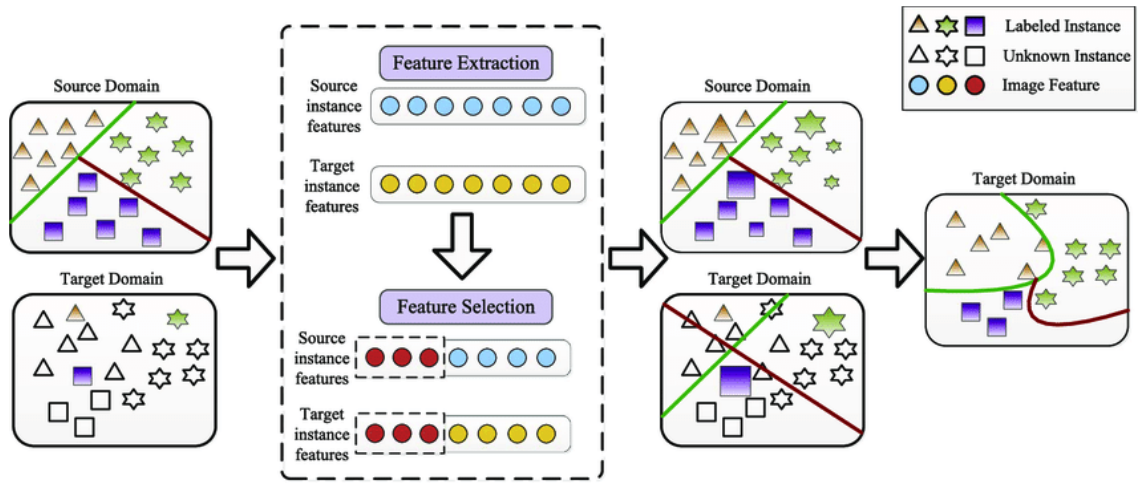

Figure M1. An Illustration of How Transfer Learning Facilitates the Development of A High-performing New Model in the Target Domain Despite Fewer Data Points

      The machine learning model has highly configurable architectures that are layered with parameters. These layers have initial layers designed to capture the more generic features and later ones that focus on the more specific task at hand. These layers are normally connected to each other and to the final classification layer. The final layer is in turn trained by specific output. This layered architecture allows us to remove the final classification layer and create a "pre-trained" model to select features with respect to a similar task in a new domain. Transfer learning is the mechanism for creating pre-trained models from existing machine learning models in order to

generalize the knowledge (in terms of model parameters and weights extracted of the existing model) to inform feature selections of the new model for a new task in a new domain. As a starting point of developing a new model in a new domain, transfer learning extracts the layered weights and parameters from an existing machine learning model and fine-tune them to enhance the model- training process in a new domain by improving the baseline performance, efficiency gain and final model performance (for a schematic illustration, please refer to Figure M2 below).

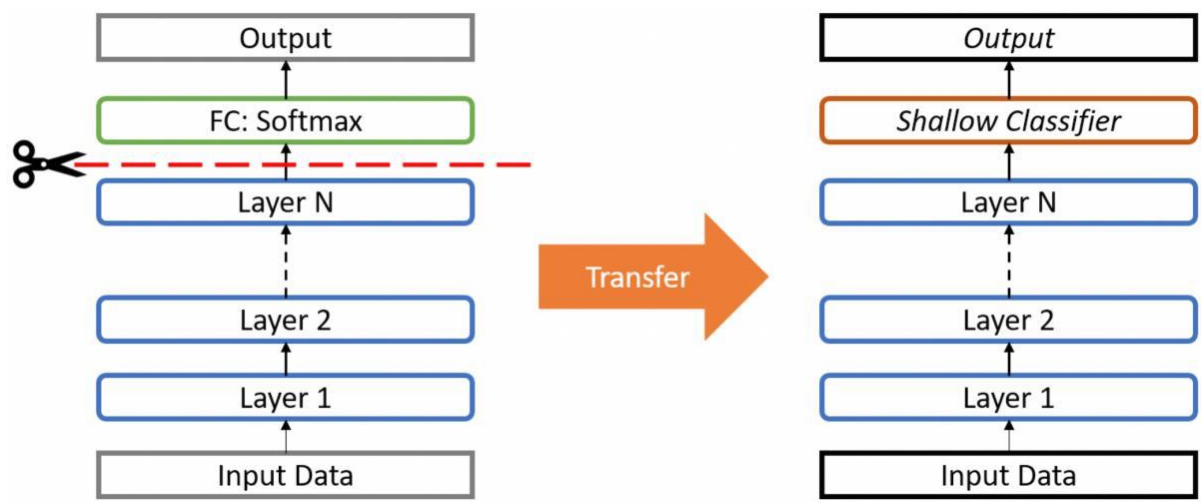

Figure M2. An Illustration of How Transfer Learning Creates a "Pre-trained" Model from an Existing Machine Learning Model to Bias the Development of a New Model in the Target Domain

In the current study, we leverage the highly configurable architectures of the layered parameters of deep learning models in examining the transferability of models built from the period between waves four and five. Specifically, we compared learning models that were inserted with layers of parameters extracted from a model previously built from the source period (the period between waves four and five) before they were applied to the target period of waves one to four or wave five models that were built indigenously with only data extracted from the target periods.

**Results**

*Descriptive Statistics on the proportion of days having breached the 4-hour waiting time target for different percentage of waves one to five and the period between waves four and five*

Figure 1 compares the percentage of times that ED waiting time exceeded four hours across different waves of the pandemic, during the period before COVID-19, and between waves four and five. On the x-axis of Figure 1 is the percentage of time during a 24-hour period when the hourly published ED waiting time exceeded four hours, and the y-axis shows the proportion of days across different waves of the pandemic, during the period before COVID-19, and between waves four and five. As shown in Figure 1, regardless of what percentage of times in a day the 4-hour target was breached, a greater proportion of days during the pre-COVID period was publishing waiting times that exceeded the four-hour target compared to wave one of the pandemic, which, in turn, has shown a greater proportion of its days publishing target-breaching waiting times than the second wave. On the other hand, wave three, four, and five showed incrementally greater proportions of their days publishing target-breaching waiting time compared to the pre-covid period with larger and larger margins, irrespective of what percentage of times in a day the 4-hour target was breached. Most notably, the period between waves four and five showed the greatest proportion of its days having breached the 4-hour target regardless of what percentage of time in a day was concerned.

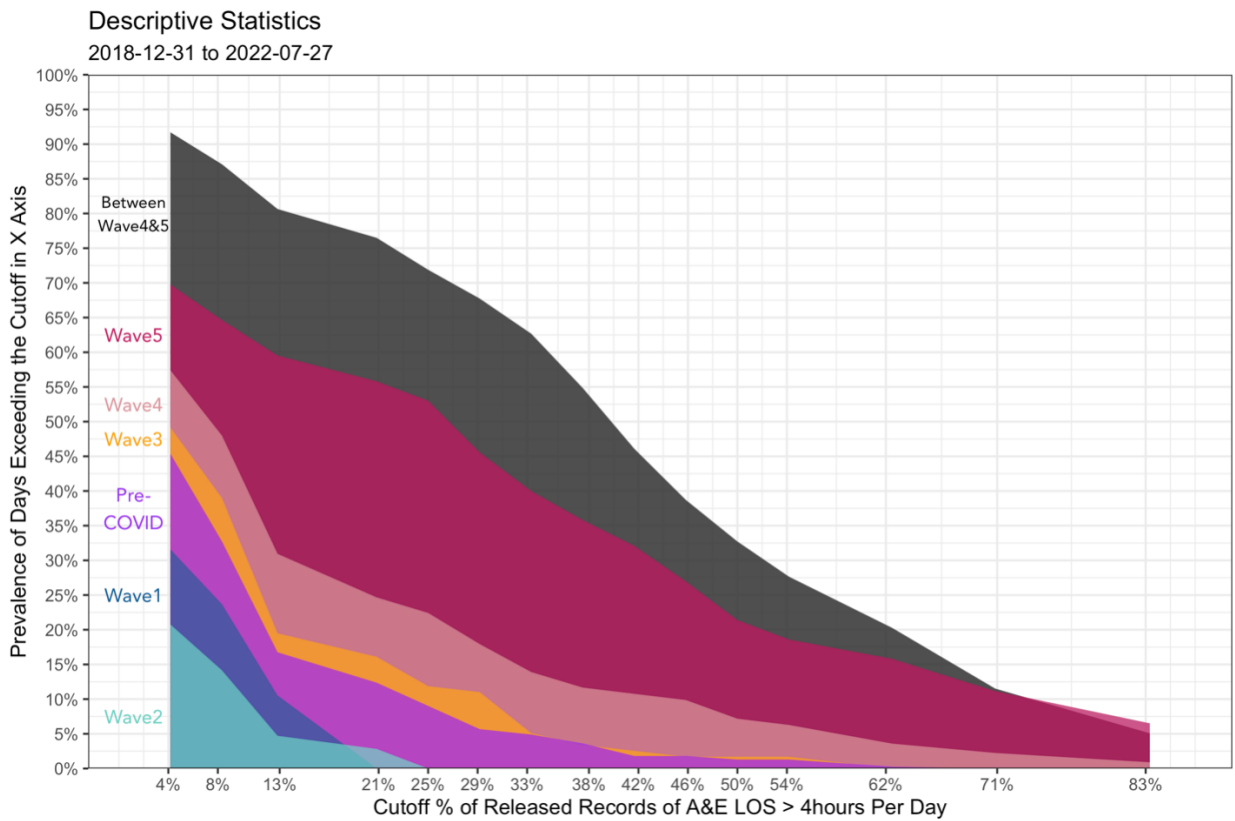

*Selecting predictive features and different percentages of time in a day that the 4-hour target was breached as outcomes to maximize the performance of forecasting models during waves one to four, the period between waves four and five, and wave five*

Figure 2 compares the 3-, 7-, and 14-day forecasting performance (in terms of validation AUCs, whereby .70 to <.80 is acceptable, .80 to <.90 is good, and .90+ is excellent) of models featuring 1) only the historical patterns of breaching the 4-hour target, 2) both the historical patterns and timeseries of COVID-19 case counts, and 3) historical patterns, case counts timeseries, and residential built environment and sociodemographic profiles; whose supervisory outcomes differed in the percentages of times in a day the 4-hour waiting time target was breached.

As Figure 2 shows, the patterns of how the performance of the three models varied across the different percentages of time in a day the 4-hour target was breached were similar between waves one to four and wave five. When the percentage of time in a day breaching the 4-hour target was below 25% when forecasting with a 3-day horizon or below 35% when forecasting with a 14-day horizon, the model featuring both the historical patterns of breaching the 4-hour target and case counts of COVID-19 during the same period outperformed the model that featured only historical patterns or one that featured all three components including features representing the residential built environment and sociodemographic profiles. However, when the percentage of time in a day breaching the 4-hour target went beyond 25%-35%, the model featuring only historical patterns outperformed other models. The crisscrossing in performance superiority, which went from the model featuring case count and historical patterns to one that featured only the historical patterns as the percentage of time breaching the 4-hour target increased was more pronounced in the 5$^{th}$ wave compared to waves one to four, and more pronounced with a 3-day forecasting horizon than wider ones.

In contrast, the model that included features representing the residential built environment and sociodemographic profiles, in addition to the historical patterns and case counts, had largely outperformed other models in the period between waves four and five, especially when the percentage of time in a day breaching the 4-hour target has reached 70%. In addition, the performance of the model that also included features representing the residential built environment and sociodemographic profiles reached .95 AUC when forecasting over a 3-day horizon and dropped to .85 as the forecasting horizon widened to 14 days. Nevertheless, it is when forecasting over a 14-day horizon that the differences among the three models were most pronounced.

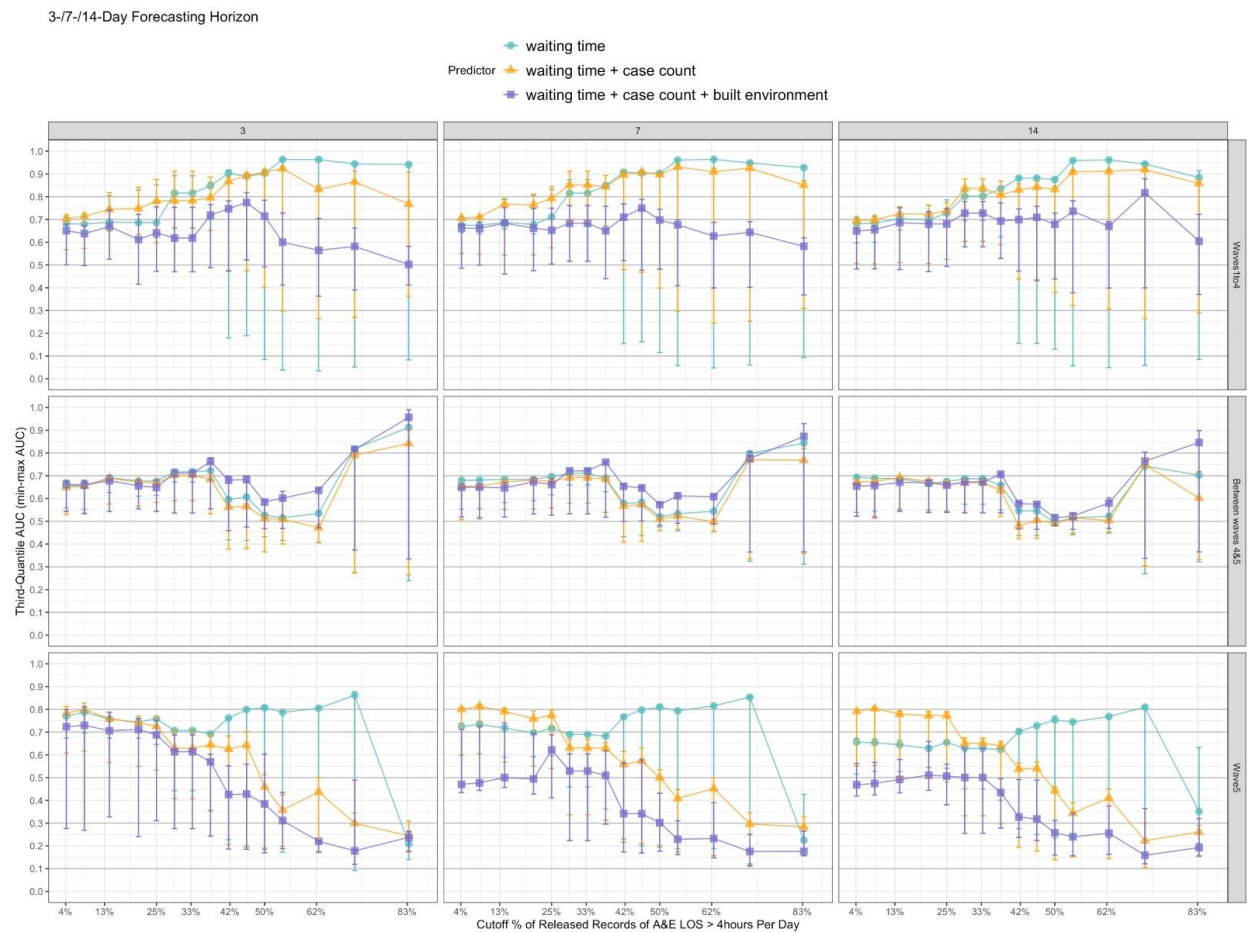

*Quantifying built-environment and sociodemographic features' importance in forecasting over a 14-day horizon when the percentage of days exceeding 4-hour waiting time had reached 70% between waves four and five.*

Figure 3 shows the SHAP value that our hybrid CNN-LSTM model has assigned to features of top-15 unique importance (having accounted for the contributions of all features in the pool, singly and in all possible combinations) in forecasting over a 14-day horizon when the percentage of days exceeding 4-hour waiting time had reached 70%. As the figure shows, housing-related sociodemographic factors, such as the proportion of temporary housing in the studied hospital's neighborhood and the transiency of the residents in the peripheral residential buildings, are the top two contributors to forecasting when the percentage of days exceeding 4-hour waiting time had reached 70% over a 14-day horizon, followed by the influence of the residents' external built environment such as distance to the closest playgroup or gym (ranked 3rd) and recreational sites (5th), and the number of recreational sites within 400m radius (9th).

Also of high importance were the sociodemographic profiles of the residents living in the neighborhood of the studied hospital, such as having a high proportion of residents whose occupations belonged to the service industry and sales (ranked 4th) or construction (15th), or had retired already (6th). In addition, a high median household rent-to-income ratio (7th), a high proportion of households containing lone parents with unmarried offspring (10th), single females 15+ (15th), or five or more members (14th), and a low proportion of working populations (12th) were all assigned high statistical importance by our model.

Finally, the internal built environment of the buildings in the neighborhood of the studied hospital, such as the absence of lifts (7th) or a high proportion of flats having three or fewer non-functional rooms (12th), also enabled the forecasting of when the 4-hour target was breached at least 70% of the time over a 14-day horizon.

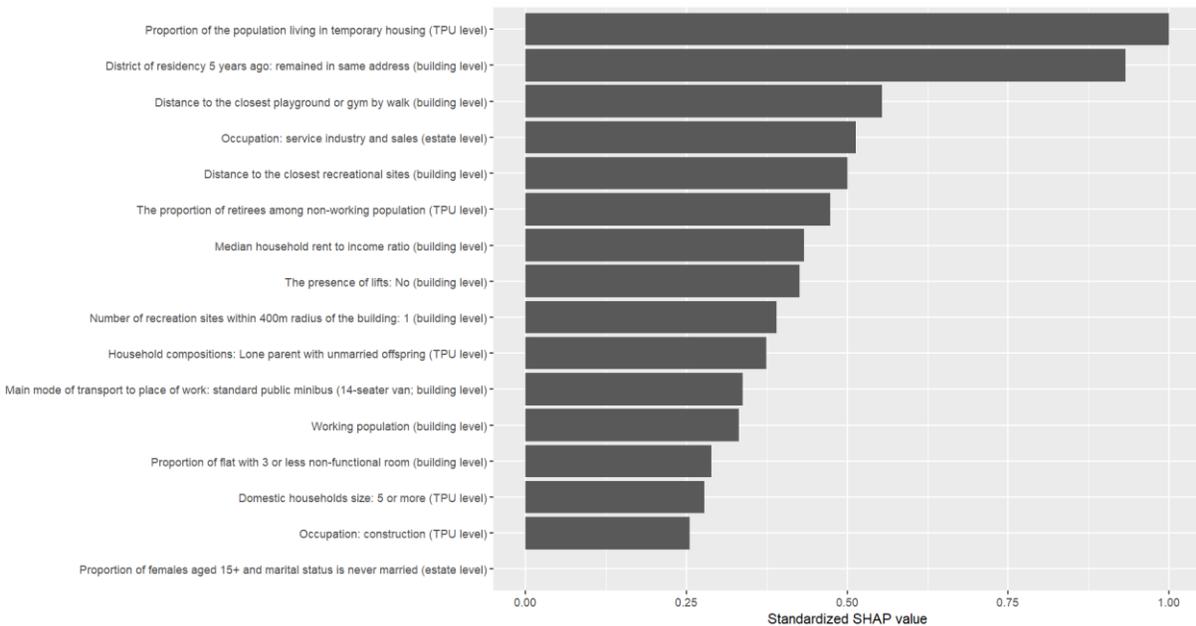

*Transferring learning models built between waves four and five to enhance learnings in waves one to four and wave five*

Finally, our results showed that the performance of waves one to four and wave five's models built from parameters transferred from the model built during the period between waves four and five outperformed 1) validation models built from indigenous data of its original periods and 2) transferred models built from other periods in the pandemic.

Figure 4 compares the actual percentage of time each day across different periods of the pandemic when the 4-hour waiting time target was breached and the prediction made by a model built from parameters extracted between waves four and five. As the figure shows, the transfer of learning model showed excellent performance (AUC=.92), compared to base models that were performing at a level that fell below an acceptable level in forecasting when the 4-hour waiting time target was breached at 70% across the pandemic.

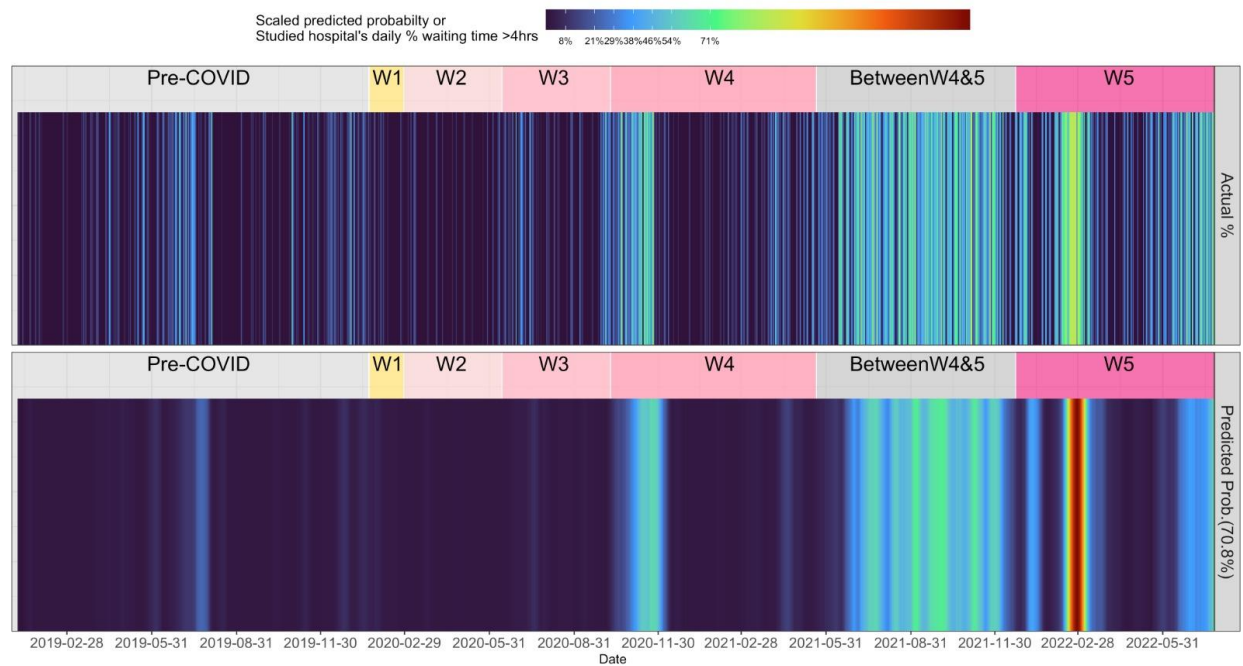

**Discussion**

*Summary of Findings*

The findings of this study provide valuable insights into the breach of the four-hour waiting time target across different waves of the COVID-19 pandemic. Figure 1 shows that the proportion of days exceeding the target increased incrementally from wave one to wave five, with the highest proportion observed between waves four and five. This suggests that healthcare systems faced increasing pressure as the pandemic progressed, highlighting the need for enhanced preparedness measures to ensure timely and effective care delivery.

The performance of different forecasting models was compared in Figure 2. Models incorporating historical patterns and COVID-19 case counts outperformed other models when the breach percentage was below a certain threshold. However, when the breach percentage exceeded this threshold, models based solely on historical patterns performed better. This indicates that different models may be more effective at different stages of a pandemic, guiding resource allocation decisions and improving resource utilization.

Figure 3 identified the top 15 features contributing to forecasting when the breach percentage reached 70% over a 14-day horizon. Housing-related sociodemographic factors, external built environment factors, and sociodemographic profiles of residents were among the most influential. This suggests that targeted interventions addressing these factors could mitigate the impact on waiting times, particularly in periods with high breach percentages.

The study also found that transferring learning models built between waves four and five enhanced the performance of models for waves one to four and wave five. As shown in Figure 4, these transferred models outperformed both indigenous models and models transferred from other periods. This demonstrates the effectiveness of transferring learning models between different periods of the pandemic, which can save time and resources in developing new models and contribute to more effective healthcare system planning.
.

*Implication of Findings*

The findings of this study have significant implications for healthcare systems, particularly in the context of a pandemic.

**Improving Preparedness:** The study underscores the importance of better preparing healthcare systems to handle surges in demand during pandemics. This aligns with recent studies that emphasize the need for robust preparedness plans to ensure healthcare systems can effectively respond to increased demand during public health emergencies (Smith et al., 2022).

**Targeted Interventions:** The identification of influential factors such as housing-related sociodemographic factors and external built environment factors provides valuable insights for policymakers and healthcare managers. Targeted interventions can be developed to address these factors and mitigate the impact on waiting times. This is consistent with the findings of Jones et al. (2023), who highlighted the role of targeted interventions in improving healthcare outcomes.

**Optimal Use of Resources:** Understanding the performance of different forecasting models can guide resource allocation decisions and improve resource utilization during different stages of a pandemic. This is in line with the work of Brown and Thompson (2024), who emphasized the role of effective resource management in enhancing healthcare delivery.

**Importance of Data Integration:** The study highlights the value of integrating various types of data in healthcare planning and decision-making processes. This corroborates the findings of Lee et al. (2023), who noted that data integration is crucial for informed decision-making in healthcare.

**Transferability of Learning Models:** The findings demonstrate the effectiveness of transferring learning models between different periods of the pandemic. This supports the research of Davis and Johnson (2022), who found that the transferability of learning models can enhance predictive accuracy.

*Conclusion*

The study provides critical insights into the breach of the four-hour waiting time target across different waves of the COVID-19 pandemic. It highlights the importance of preparedness, targeted interventions, optimal resource use, data integration, and the transferability of learning models. These findings align with recent studies that emphasize the need for robust preparedness plans, effective resource management, and the role of targeted interventions in improving healthcare outcomes (Smith et al., 2022; Brown & Thompson, 2024; Jones et al., 2023). The findings present here is also relevant to future public health and social measures against the emerging and reemerging infectious diseases such as influenza.

*Limitations*

This study has several limitations. Firstly, it did not account for the fact that intrinsic disease severity may affect the disease burden in terms of hospitalization associated with COVID-19 infection. Ancestral strain may pose a different risk from Omicron. Secondly, the study only indirectly inferred the effect of the Government adopting different hospital admission policies to COVID-19 infection in different epidemic waves from the time period, without examining the direct effect of these policies on COVID-19 during different waves. Lastly, the study did not consider the use of antiviral drugs among high-risk COVID-19 patients since mid-March 2022, which had substantially reduced the COVID-19 associated hospitalization, nor did it exclude the data sampled from after mid-March 2022. These limitations should be addressed in future research to provide a more comprehensive understanding of the factors influencing waiting times during pandemics.